\documentclass[letterpaper, 10 pt, conference]{ieeeconf}  

\IEEEoverridecommandlockouts                              

\usepackage{amsmath}
\usepackage{booktabs}
\usepackage{tabularx}
\usepackage{graphicx}
\usepackage{xcolor} 
\usepackage{multirow}
\usepackage{cite}

\title{\LARGE \bf
Neighbor-Aware Reinforcement Learning for Mixed Traffic Optimization in Large-scale Networks
}

\author{Iftekharul Islam and Weizi Li 
\thanks{Iftekharul Islam and Weizi Li are with Min H. Kao Department of Electrical Engineering and Computer Science at University of Tennessee, Knoxville, TN, USA {\tt\small mislam73@vols.utk.edu, weizili@utk.edu}}%
}

\begin{document}

\maketitle
\thispagestyle{empty}
\pagestyle{empty}

\begin{abstract}

Managing mixed traffic comprising human-driven and robot vehicles (RVs) across large-scale networks presents unique challenges beyond single-intersection control. This paper proposes a reinforcement learning framework for coordinating mixed traffic across multiple interconnected intersections. Our key contribution is a neighbor-aware reward mechanism that enables RVs to maintain balanced distribution across the network while optimizing local intersection efficiency. We evaluate our approach using a real-world network, demonstrating its effectiveness in managing realistic traffic patterns. Results show that our method reduces average waiting times by 39.2\% compared to the state-of-the-art single-intersection control policy and 79.8\% compared to traditional traffic signals. The framework's ability to coordinate traffic across multiple intersections while maintaining balanced RV distribution provides a foundation for deploying learning-based solutions in urban traffic systems.

\end{abstract}


\section{INTRODUCTION}
\label{introduction}


Traffic congestion in urban environments continues to pose significant challenges to mobility, economic growth, and quality of life. Despite the implementation of various traffic control methods, congestion remains a global issue. The issue is particularly acute in urban areas, where drivers lose up to 101 hours annually, resulting in economic losses exceeding \$9 billion each year due to wasted time~\cite{INRIX2023}. Modern urban road networks primarily consist of linearly-coupled roads interconnected by intersections, making intersections critical points for traffic management and potential bottlenecks in the system.

The emergence of autonomous vehicles presents new opportunities for traffic control and optimization. However, the transition to full autonomy will be gradual, leading to a prolonged period of mixed traffic where human-driven vehicles (HVs) and robot vehicles (RVs) coexist. Recent studies have demonstrated that RVs can effectively influence surrounding traffic flow and improve overall system efficiency~\cite{gholamhosseinian2022comprehensive, qadri2020state}. In particular, researchers have shown that RVs can successfully coordinate traffic at individual intersections without traditional traffic signals, achieving significant reductions in average waiting times and greenhouse gas emissions~\cite{Islam2024Heterogeneous,Wang2024Intersection,Wang2024Privacy,Poudel2024CARL,Poudel2024EnduRL,Villarreal2024Eco,Villarreal2023Pixel,Villarreal2023Chat,seong2021learning}.

Yet, scaling these intersection-level control strategies to large networks presents several critical challenges. First, maintaining balanced RV distribution across multiple intersections is essential for effective traffic control but becomes increasingly complex in large-scale deployments. Second, decisions made at one intersection can significantly impact neighboring intersections, requiring coordination mechanisms that consider network-wide effects. Third, the dynamic and stochastic nature of urban traffic demands adaptive control strategies that can respond to changing conditions while ensuring stable system performance.

While previous work has explored various aspects of mixed traffic control, including intersection management~\cite{dresner2008multiagent, jin2013platoon}, vehicle coordination~\cite{mirheli2019consensus, malikopoulos2018decentralized}, and reinforcement learning-based control~\cite{yan2021reinforcement, peng2021connected}, the challenge of coordinating mixed traffic across large-scale networks remains largely unaddressed. Existing approaches either focus on single intersections or assume full autonomy, limiting their applicability to real-world mixed traffic scenarios.

This paper presents a reinforcement learning framework for controlling mixed traffic across large-scale networks. Our key contributions include:
\begin{itemize}
    \item A network-aware control framework that enables efficient coordination of mixed traffic across multiple interconnected intersections.
    \item A novel neighbor-aware reward mechanism that maintains balanced RV distribution while optimizing local intersection efficiency.
    \item Comprehensive evaluation using a real-world network from Colorado Springs, demonstrating significant improvements over existing methods.
\end{itemize}

We evaluate our approach using a network of 17 intersections, demonstrating its effectiveness in managing realistic traffic patterns. Results show that our method reduces average waiting times by 39.2\% compared to the state-of-the-art single-intersection control policy and 79.8\% compared to traditional traffic signals. These improvements stem from our framework's ability to balance local intersection efficiency with network-wide coordination, providing a foundation for deploying learning-based solutions in urban traffic systems.

\section{RELATED WORK}
\label{related_work}

Urban intersection management has become increasingly complex with the growing coexistence of autonomous and human-driven vehicles. Traditional traffic control methods, such as fixed-time and actuated signals, have been widely used to manage traffic flow in predictable conditions but often fail to adapt to dynamic and complex environments~\cite{gholamhosseinian2022comprehensive}. While optimization-based approaches, including integer linear programming and rule-based strategies, offer enhanced efficiency, they face significant scalability challenges in large-scale networks~\cite{qadri2020state}.

The management of traffic at unsignalized intersections has been extensively studied, particularly for connected autonomous vehicles (CAVs). Dresner and Stone~\cite{dresner2008multiagent} introduced a multi-agent reservation-based mechanism where CAVs secure time-space slots in a first-come, first-serve manner. Jin et al.~\cite{jin2013platoon} extended this concept to manage CAV platoons, enabling more efficient intersection scheduling. Other methods, such as the controllable gap strategy by Chen et al.~\cite{chen2022improved}, dynamically adjust inter-vehicle gaps to minimize collision risks. 
Decentralized approaches have also gained attention, with methods such as consensus-based trajectory control~\cite{mirheli2019consensus} and energy-optimized vehicle coordination~\cite{malikopoulos2018decentralized}. However, these studies predominantly focus on fully autonomous environments, limiting their application to real-world mixed traffic scenarios.

Reinforcement learning (RL) has emerged as a promising tool for optimizing mixed traffic control. Wang et al.~\cite{Wang2024Intersection,Wang2024Privacy} employed decentralized multi-agent RL to manage mixed traffic at unsignalized intersections, achieving significant reductions in average waiting times. Similarly, Yan and Wu~\cite{yan2021reinforcement} proposed a model-free RL framework for controlling mixed traffic, while Peng et al.~\cite{peng2021connected} demonstrated improvements in traffic flow and conflict reduction through RL-based CAV coordination. Villarreal et al.~\cite{Villarreal2023Pixel} used image-based observations for decision-making. Poudel et al.~\cite{Poudel2024CARL} incorporated real-world driving profiles to enhance traffic stability and safety. Shi et al.~\cite{shi2022control} further demonstrated RL’s ability to outperform traditional control methods in urban unsignalized intersections.

Scaling RL-based traffic control to large networks has been a key focus of recent research. Zhao et al.\cite{ma2024efficient} proposed a decentralized RL framework for large-scale traffic signal control, employing a multi-agent approach that ensures scalability through localized policy optimization and indirect collaboration via shared state information. Wang et al.\cite{Wang2024Privacy} extended RL-based coordination to city-scale mixed traffic networks, integrating dynamic vehicle routing and privacy-preserving crowdsourcing to balance RV penetration rates and prevent localized congestion. 

This study proposes a RL-based mixed traffic control approach designed to address the challenges of large-scale network. We introduce a neighbor-aware reward mechanism to balance robot vehicle (RV) distributions and improve coordination across interconnected intersections. Unlike previous approaches, the framework prioritizes both local intersection efficiency and network-wide traffic balance, demonstrating significant potential for managing real-world traffic systems with mixed autonomy. By addressing critical gaps in scalability and coordination, this work provides a foundation for deploying RL-based solutions in urban traffic networks.

\section{METHODOLOGY}
\label{methodology}

\subsection{RL-based Network-Level Control Framework}
The control of large-scale mixed traffic presents unique challenges not addressed by single-intersection approaches:

1. \textbf{Local Control}: Each RV must make efficient stop/go decisions at individual intersections.
2. \textbf{Network Coordination}: RVs need to maintain balanced distribution across the network.
3. \textbf{Downstream Impact}: Decisions at one intersection affect traffic conditions at neighboring intersections.

We formulate this as a POMDP defined by the tuple $(\mathcal{S}, \mathcal{A}, \mathcal{T}, \mathcal{R}, \mathcal{O}, \mathcal{Z}, \gamma)$, where:
\begin{itemize}
    \item $\mathcal{S}$: State space representing traffic conditions across the entire network
    \item $\mathcal{A} = \{\text{Stop}, \text{Go}\}$: Action space for RVs
    \item $\mathcal{T}(s'|s,a)$: Transition probability between states
    \item $\mathcal{R}$: Our novel neighbor-aware reward function
    \item $\mathcal{O}$: Observation space for each RV
    \item $\mathcal{Z}(o|s)$: Observation probability function
    \item $\gamma$: Discount factor
\end{itemize}

\begin{figure}[b]
    \centering
    \includegraphics[width=\columnwidth]{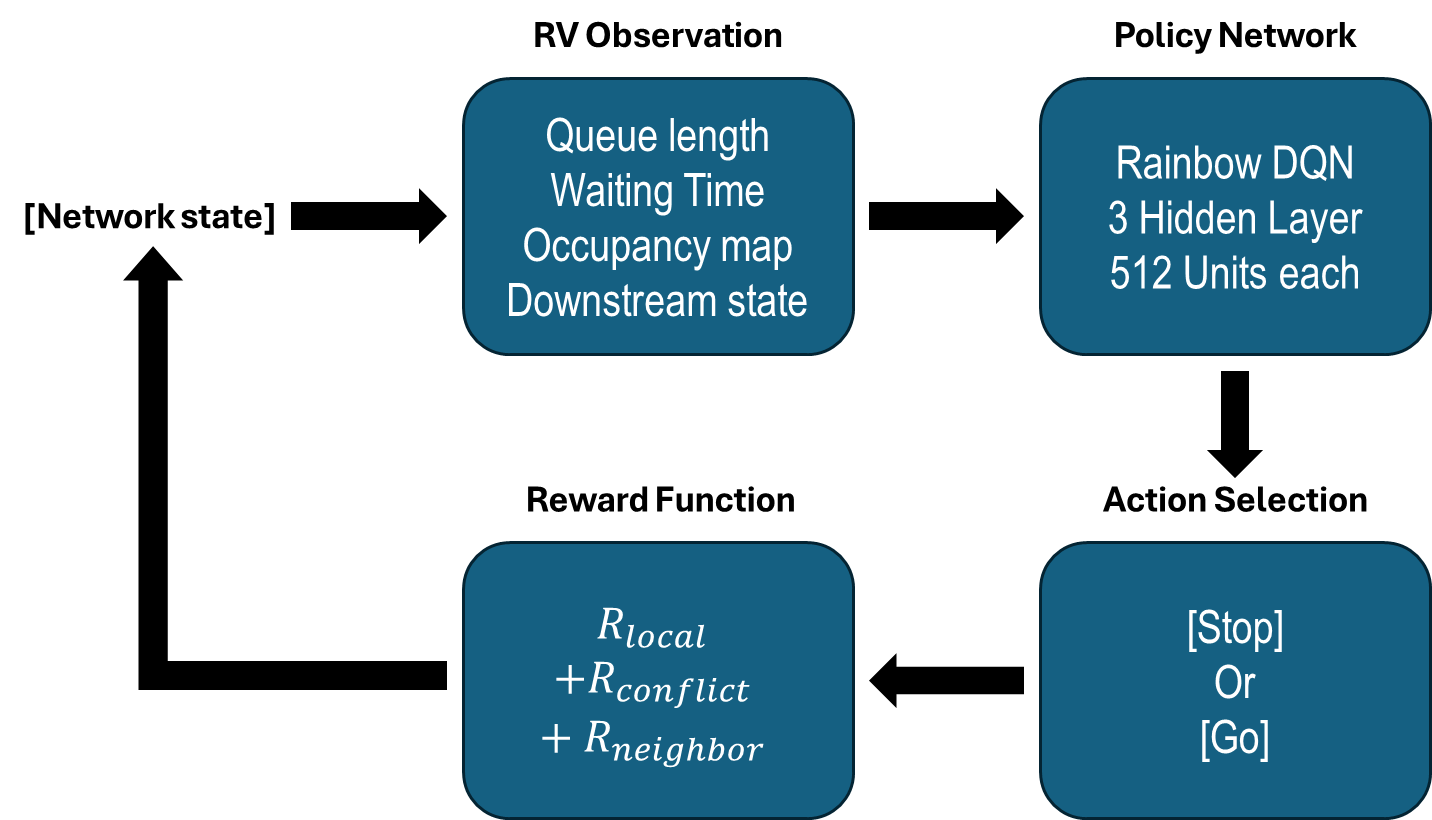}
    \caption{Overview of the neighbor-aware reinforcement learning framework for large-scale mixed traffic control. The system operates in a continuous loop where each RV observes the network state (including queue length, waiting time, occupancy map, and downstream conditions), processes this information through a Rainbow DQN policy network (three hidden layers with 512 units each), and selects either Stop or Go actions. The reward function combines three components: local efficiency ($R_{local}$), conflict avoidance ($R_{conflict}$), and neighbor-aware distribution ($R_{neighbor}$). This feedback loop enables the system to learn policies that balance local intersection efficiency with network-wide RV distribution.}
    \label{fig:methodology}
\end{figure}

Fig.~\ref{fig:methodology} shows the key component of the RL framework.

\subsection{Network-Aware State Representation}
At each timestep $t$, an RV observes both local intersection conditions and downstream network state:
\[
y_t = \bigoplus_{d \in \mathcal{D}} \langle q_d, w_d, o_d \rangle \oplus s_{\text{next}},
\]
where $\mathcal{D}$ represents incoming traffic directions, $q_d$ is queue length, $w_d$ is average waiting time, $o_d$ indicates intersection occupancy, and $s_{\text{next}}$ represents the state of the downstream intersection.

\subsection{Neighbor-Aware Reward Design}
Our key contribution is a novel reward function that balances local efficiency with network-wide RV distribution:
\[
R(s,a) = R_{\text{local}}(s,a) + R_{\text{conflict}}(s,a) + \alpha R_{\text{neighbor}}(s,a)
\]
where:
\begin{itemize}
    \item $R_{\text{local}}$ encourages efficient local traffic flow:
    \[
    R_{\text{local}}(s,a) = 
    \begin{cases}
    -w_d & \text{if } a = \text{Stop}, \\
    w_d & \text{if } a = \text{Go}.
    \end{cases}
    \]
    \item $R_{\text{conflict}}$ penalizes potential conflicts:
    \[
    R_{\text{conflict}}(s,a) = 
    \begin{cases}
    -1 & \text{if conflict occurs}, \\
    0 & \text{otherwise}.
    \end{cases}
    \]
    \item $R_{\text{neighbor}}$ promotes balanced RV distribution:
    \[
    R_{\text{neighbor}}(s,a) = 
    \begin{cases}
    \max(0, p_{\text{target}} - p_{\text{current}}) & \text{if } a = \text{Go}, \\
    0 & \text{if } a = \text{Stop}.
    \end{cases}
    \]
\end{itemize}

\subsection{Large-Scale Network Architecture}
\begin{figure}[b]
    \centering
    \includegraphics[width=1.0\columnwidth]{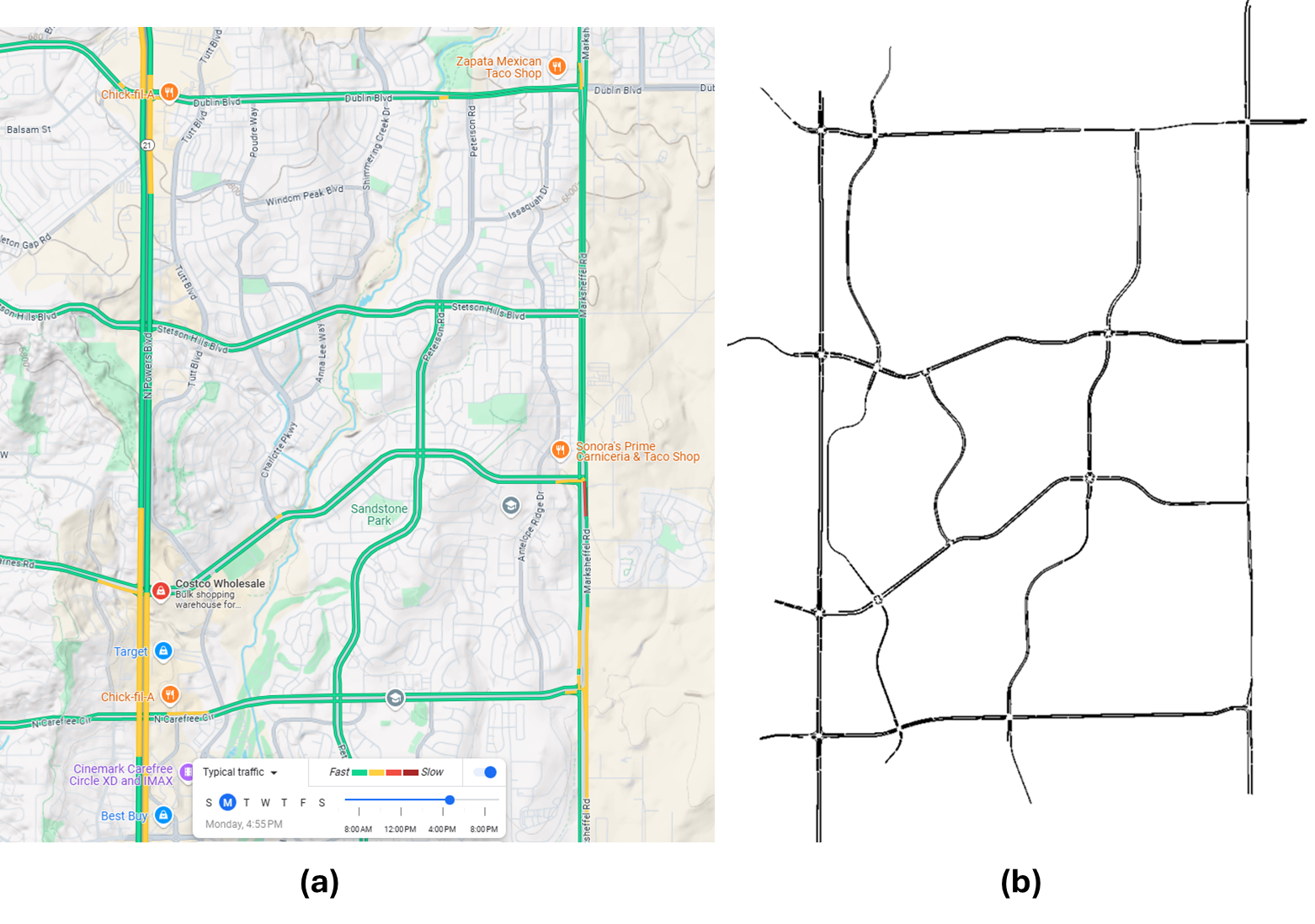} 
    \caption{Network representation of the study area in Colorado Springs, CO, USA. (a) Google Maps visualization showing typical traffic conditions during Monday evening rush hour (4:55 PM).
    (b) Simplified network after removing secondary roads.}
    \label{fig:colorado_network}
\end{figure}

Our study area is a real-world urban road network from Colorado Springs, CO, USA, consisting of multiple interconnected major intersections. The network layout was extracted from the city's primary road infrastructure, focusing on major arterial roads and excluding secondary streets to capture key traffic flow patterns. The resulting network comprises 17 intersections of varying configurations (both four-way and three-way), representing the core traffic corridors in the area (Fig.~\ref{fig:colorado_network}).

Traffic demand is modeled based on typical patterns observed on Google Maps, ensuring the simulations reflect realistic urban conditions. These patterns are implemented to create diverse and representative vehicle routes across the network.

Each intersection in our network maintains a control zone extending 30 meters from its center, where robot vehicles (RVs) observe and react to traffic conditions. This real-world setup provides a robust testbed for evaluating our control strategy under realistic conditions, including varying intersection geometries, road capacities, and traffic patterns typically observed during peak hours.

\section{Experiments and Results}
\label{experiments}

\subsection{Experimental Setup}
We evaluate our framework using the large-scale traffic network in Colorado Springs, CO, USA. 
The RL policy is implemented using the Rainbow DQN algorithm~\cite{hessel2018rainbow} with a three-layer neural network architecture, where each hidden layer contains 512 units. The network is trained with a learning rate of 0.0005 and a discount factor of 0.99. Training is conducted on an Intel i9-13900KF CPU and an NVIDIA GeForce RTX 4090 GPU, running for 1,000 iterations to ensure policy convergence.
To ensure robust evaluation of our approach, we conduct three independent evaluation runs. Each evaluation run:
\begin{itemize}
    \item lasts 1,500 simulation seconds,
    \item uses the last 1,000 seconds for waiting time computation, allowing initial 500 seconds for system stabilization, and,
    \item maintains a fixed RV penetration rate of 60\%, chosen to balance the benefits of autonomous control with practical deployment considerations.
\end{itemize}

To evaluate the performance of our approach, we compare against three baselines:

\begin{enumerate}
    \item \textbf{HV-Signalized:} All human-driven vehicles (HVs) with traditional pre-timed traffic signals, representing the standard method in urban traffic management.
    \item \textbf{HV-Unsignalized:} All HVs without traffic signals, representing uncontrolled traffic behavior at intersections.
    \item \textbf{Local-RL:} The mixed traffic coordination method by Wang et al.~\cite{Wang2024Intersection}, designed for single intersections, used here to highlight the benefits of our network-level coordination approach.
\end{enumerate}

To assess performance, we focus on average waiting time as our primary metric, defined as the duration vehicles remain stationary in intersection control zones. This metric effectively captures both local intersection efficiency and network-level coordination effects. Additionally, we examine RV distribution patterns across intersections to evaluate the effectiveness of our neighbor-aware routing mechanism.

\subsection{Overall Performance}
Table~\ref{tab:overall_results} presents the average waiting times across different control strategies:

\begin{table}[h]
\centering
\caption{Performance Comparison Across Control Strategies. Results show average waiting times measured on the 17-intersection network. Our Neighbor-Aware Policy with 60\% RV penetration rate achieves a 39.2\% reduction compared to Local-RL (single-intersection baseline) and a 79.8\% reduction compared to HV-Signalized (traditional traffic signals). HV-Unsignalized demonstrates the baseline scenario without any traffic control.}
\label{tab:overall_results}
\begin{tabular}{lccc}
\hline
Control Strategy & Average Waiting Time (s) \\
\hline
HV-Signalized & 10.60 \\
HV-Unsignalized & 22.33 \\
Local-RL & 3.52 \\
Our Neighbor-Aware Policy & 2.14 \\
\hline
\end{tabular}
\end{table}

Our experiments demonstrate the effectiveness of intelligent traffic control in large-scale networks. The HV-Unsignalized scenario, with an average waiting time of 22.33 seconds, highlights the fundamental need for traffic management at complex intersections. Traditional signal control (HV-Signalized) improves this baseline significantly to 10.60 seconds, while learning-based approaches achieve substantially better performance. With a 60\% RV penetration rate, our neighbor-aware policy achieves an average waiting time of 2.14 seconds, representing a 39.2\% reduction compared to Local-RL (3.52 seconds) and a 79.8\% reduction compared to HV-Signalized. These results demonstrate that network-level coordination through our neighbor-aware mechanism provides significant advantages over both traditional methods and single-intersection RL approach.

The significant performance improvements observed with our neighbor-aware policy likely stem from its ability to consider both local intersection efficiency and network-wide RV distribution. Unlike the original policy that makes decisions based solely on local intersection conditions, our approach enables RVs to consider downstream traffic states and adjust their behavior accordingly. This network-level awareness helps maintain more balanced RV distribution across intersections, potentially preventing the formation of bottlenecks that could arise from RV-sparse regions. 

\section{CONCLUSION AND FUTURE WORK}
\label{conclusion}

This study presents a scalable reinforcement learning (RL)-based framework for managing traffic in large-scale urban networks. By introducing a novel neighbor-aware reward mechanism, the framework effectively balances robot vehicle (RV) distributions across interconnected intersections, improving both local and network-wide traffic efficiency. Simulations on a 17-intersection network demonstrate significant reductions in average waiting times compared to traditional traffic management methods and single-intersection RL-based approaches. These results validate the feasibility of deploying RL-based traffic management systems in real-world urban networks with mixed traffic scenarios.


We see several promising directions for future work. 
First, we aim to integrate the insights from this project with existing techniques~\cite{Li2019ADAPS,Shen2022IRL,Lin2022Attention,Shen2021Corruption,Villarreal2024AutoJoin} to enhance the learning, planning, and robustness capabilities of robotic vehicles.
Second, we plan to incorporate our controller into large-scale traffic simulation, reconstruction, and prediction frameworks~\cite{Wilkie2015Virtual,Li2017CityFlowRecon,Li2017CitySparseITSM,Li2018CityEstIET,Lin2019Compress,Lin2019BikeTRB,Chao2020Survey,Poudel2021Attack,Lin2022GCGRNN,Guo2024Simulation}, aiming to support a wide range of applications in intelligent transportation systems.
Finally, we intend to test our approach on hardware~\cite{Poudel2022Micro} and explore its integration with inter-vehicle communication through network optimization~\cite{Wickman2022SparRL}.


\bibliographystyle{./IEEEtran} 
\bibliography{./references}

\begin{thebibliography}{10}
\providecommand{\url}[1]{#1}
\csname url@rmstyle\endcsname
\providecommand{\newblock}{\relax}
\providecommand{\bibinfo}[2]{#2}
\providecommand\BIBentrySTDinterwordspacing{\spaceskip=0pt\relax}
\providecommand\BIBentryALTinterwordstretchfactor{4}
\providecommand\BIBentryALTinterwordspacing{\spaceskip=\fontdimen2\font plus
\BIBentryALTinterwordstretchfactor\fontdimen3\font minus \fontdimen4\font\relax}
\providecommand\BIBforeignlanguage[2]{{%
\expandafter\ifx\csname l@#1\endcsname\relax
\typeout{** WARNING: IEEEtran.bst: No hyphenation pattern has been}%
\typeout{** loaded for the language `#1'. Using the pattern for}%
\typeout{** the default language instead.}%
\else
\language=\csname l@#1\endcsname
\fi
#2}}

\bibitem{INRIX2023}
\BIBentryALTinterwordspacing
INRIX, ``2023 global traffic scorecard,'' 2024, accessed December 1, 2024. [Online]. Available: \url{https://inrix.com/scorecard/}
\BIBentrySTDinterwordspacing

\bibitem{gholamhosseinian2022comprehensive}
A.~Gholamhosseinian and J.~Seitz, ``A comprehensive survey on cooperative intersection management for heterogeneous connected vehicles,'' \emph{IEEE Access}, vol.~10, pp. 7937--7972, 2022.

\bibitem{qadri2020state}
S.~S. S.~M. Qadri, M.~A. G{\"o}k{\c{c}}e, and E.~{\"O}ner, ``State-of-art review of traffic signal control methods: challenges and opportunities,'' \emph{European transport research review}, vol.~12, pp. 1--23, 2020.

\bibitem{Islam2024Heterogeneous}
I.~Islam, W.~Li, S.~Li, and K.~Heaslip, ``Heterogeneous mixed traffic control and coordination,'' in \emph{Heterogeneous Mixed Traffic Control and Coordination}, 2024.

\bibitem{Wang2024Intersection}
D.~Wang, W.~Li, L.~Zhu, and J.~Pan, ``Learning to control and coordinate mixed traffic through robot vehicles at complex and unsignalized intersections,'' \emph{International Journal of Robotics Research}, 2024.

\bibitem{Wang2024Privacy}
D.~Wang, W.~Li, and J.~Pan, ``Large-scale mixed traffic control using dynamic vehicle routing and privacy-preserving crowdsourcing,'' \emph{IEEE Internet of Things Journal}, vol.~11, no.~2, pp. 1981--1989, 2024.

\bibitem{Poudel2024CARL}
B.~Poudel, W.~Li, and S.~Li, ``Carl: Congestion-aware reinforcement learning for imitation-based perturbations in mixed traffic control,'' in \emph{International Conference on CYBER Technology in Automation, Control, and Intelligent Systems (CYBER)}, 2024.

\bibitem{Poudel2024EnduRL}
B.~Poudel, W.~Li, and K.~Heaslip, ``Endurl: Enhancing safety, stability, and efficiency of mixed traffic under real-world perturbations via reinforcement learning,'' in \emph{IEEE/RSJ International Conference on Intelligent Robots and Systems (IROS)}, 2024.

\bibitem{Villarreal2024Eco}
M.~Villarreal, D.~Wang, J.~Pan, and W.~Li, ``Analyzing emissions and energy efficiency in mixed traffic control at unsignalized intersections,'' in \emph{IEEE Forum for Innovative Sustainable Transportation Systems (FISTS)}, 2024, pp. 1--7.

\bibitem{Villarreal2023Pixel}
M.~Villarreal, B.~Poudel, J.~Pan, and W.~Li, ``Mixed traffic control and coordination from pixels,'' in \emph{IEEE International Conference on Robotics and Automation (ICRA)}, 2024, pp. 4488--4494.

\bibitem{Villarreal2023Chat}
M.~Villarreal, B.~Poudel, and W.~Li, ``Can chatgpt enable its? the case of mixed traffic control via reinforcement learning,'' in \emph{IEEE International Conference on Intelligent Transportation Systems (ITSC)}, 2023, pp. 3749--3755.

\bibitem{seong2021learning}
H.~Seong, C.~Jung, S.~Lee, and D.~H. Shim, ``Learning to drive at unsignalized intersections using attention-based deep reinforcement learning,'' in \emph{2021 IEEE International Intelligent Transportation Systems Conference (ITSC)}.\hskip 1em plus 0.5em minus 0.4em\relax IEEE, 2021, pp. 559--566.

\bibitem{dresner2008multiagent}
K.~Dresner and P.~Stone, ``A multiagent approach to autonomous intersection management,'' \emph{Journal of artificial intelligence research}, vol.~31, pp. 591--656, 2008.

\bibitem{jin2013platoon}
Q.~Jin, G.~Wu, K.~Boriboonsomsin, and M.~Barth, ``Platoon-based multi-agent intersection management for connected vehicle,'' in \emph{16th international ieee conference on intelligent transportation systems (itsc 2013)}.\hskip 1em plus 0.5em minus 0.4em\relax IEEE, 2013, pp. 1462--1467.

\bibitem{mirheli2019consensus}
A.~Mirheli, M.~Tajalli, L.~Hajibabai, and A.~Hajbabaie, ``A consensus-based distributed trajectory control in a signal-free intersection,'' \emph{Transportation research part C: emerging technologies}, vol. 100, pp. 161--176, 2019.

\bibitem{malikopoulos2018decentralized}
A.~A. Malikopoulos, C.~G. Cassandras, and Y.~J. Zhang, ``A decentralized energy-optimal control framework for connected automated vehicles at signal-free intersections,'' \emph{Automatica}, vol.~93, pp. 244--256, 2018.

\bibitem{yan2021reinforcement}
Z.~Yan and C.~Wu, ``Reinforcement learning for mixed autonomy intersections,'' in \emph{2021 IEEE International Intelligent Transportation Systems Conference (ITSC)}.\hskip 1em plus 0.5em minus 0.4em\relax IEEE, 2021, pp. 2089--2094.

\bibitem{peng2021connected}
B.~Peng, M.~F. Keskin, B.~Kulcs{\'a}r, and H.~Wymeersch, ``Connected autonomous vehicles for improving mixed traffic efficiency in unsignalized intersections with deep reinforcement learning,'' \emph{Communications in Transportation Research}, vol.~1, p. 100017, 2021.

\bibitem{chen2022improved}
X.~Chen, M.~Hu, B.~Xu, Y.~Bian, and H.~Qin, ``Improved reservation-based method with controllable gap strategy for vehicle coordination at non-signalized intersections,'' \emph{Physica A: Statistical Mechanics and its Applications}, vol. 604, p. 127953, 2022.

\bibitem{shi2022control}
Y.~Shi, Y.~Liu, Y.~Qi, and Q.~Han, ``A control method with reinforcement learning for urban un-signalized intersection in hybrid traffic environment,'' \emph{Sensors}, vol.~22, no.~3, p. 779, 2022.

\bibitem{ma2024efficient}
C.~Ma, A.~Li, Y.~Du, H.~Dong, and Y.~Yang, ``Efficient and scalable reinforcement learning for large-scale network control,'' \emph{Nature Machine Intelligence}, vol.~6, no.~9, pp. 1006--1020, 2024.

\bibitem{hessel2018rainbow}
M.~Hessel, J.~Modayil, H.~Van~Hasselt, T.~Schaul, G.~Ostrovski, W.~Dabney, D.~Horgan, B.~Piot, M.~Azar, and D.~Silver, ``Rainbow: Combining improvements in deep reinforcement learning,'' in \emph{Proceedings of the AAAI conference on artificial intelligence}, vol.~32, no.~1, 2018.

\bibitem{Li2019ADAPS}
W.~Li, D.~Wolinski, and M.~C. Lin, ``{ADAPS}: Autonomous driving via principled simulations,'' in \emph{IEEE International Conference on Robotics and Automation (ICRA)}, 2019, pp. 7625--7631.

\bibitem{Shen2022IRL}
Y.~Shen, W.~Li, and M.~C. Lin, ``Inverse reinforcement learning with hybrid-weight trust-region optimization and curriculum learning for autonomous maneuvering,'' in \emph{IEEE/RSJ International Conference on Intelligent Robots and Systems (IROS)}, 2022, pp. 7421--7428.

\bibitem{Lin2022Attention}
L.~Lin, W.~Li, H.~Bi, and L.~Qin, ``Vehicle trajectory prediction using {LSTM}s with spatial-temporal attention mechanisms,'' \emph{IEEE Intelligent Transportation Systems Magazine}, vol.~14, no.~2, pp. 197–--208, 2022.

\bibitem{Shen2021Corruption}
Y.~Shen, L.~Zheng, M.~Shu, W.~Li, T.~Goldstein, and M.~C. Lin, ``Gradient-free adversarial training against image corruption for learning-based steering,'' in \emph{Advances in Neural Information Processing Systems (NeurIPS)}, 2021, pp. 26\,250--26\,263.

\bibitem{Villarreal2024AutoJoin}
M.~Villarreal, B.~Poudel, R.~Wickman, Y.~Shen, and W.~Li, ``Autojoin: Efficient adversarial training for robust maneuvering via denoising autoencoder and joint learning,'' in \emph{IEEE/RSJ International Conference on Intelligent Robots and Systems (IROS)}, 2024.

\bibitem{Wilkie2015Virtual}
D.~Wilkie, J.~Sewall, W.~Li, and M.~C. Lin, ``Virtualized traffic at metropolitan scales,'' \emph{Frontiers in Robotics and AI}, vol.~2, p.~11, 2015.

\bibitem{Li2017CityFlowRecon}
W.~Li, D.~Wolinski, and M.~C. Lin, ``City-scale traffic animation using statistical learning and metamodel-based optimization,'' \emph{ACM Trans. Graph.}, vol.~36, no.~6, pp. 200:1--200:12, 2017.

\bibitem{Li2017CitySparseITSM}
W.~Li, D.~Nie, D.~Wilkie, and M.~C. Lin, ``Citywide estimation of traffic dynamics via sparse {GPS} traces,'' \emph{IEEE Intelligent Transportation Systems Magazine}, vol.~9, no.~3, pp. 100--113, 2017.

\bibitem{Li2018CityEstIET}
W.~Li, M.~Jiang, Y.~Chen, and M.~C. Lin, ``Estimating urban traffic states using iterative refinement and wardrop equilibria,'' \emph{IET Intelligent Transport Systems}, vol.~12, no.~8, pp. 875--883, 2018.

\bibitem{Lin2019Compress}
L.~Lin, W.~Li, and S.~Peeta, ``Efficient data collection and accurate travel time estimation in a connected vehicle environment via real-time compressive sensing,'' \emph{Journal of Big Data Analytics in Transportation}, vol.~1, no.~2, pp. 95--107, 2019.

\bibitem{Lin2019BikeTRB}
------, ``Predicting station-level bike-sharing demands using graph convolutional neural network,'' in \emph{Transportation Research Board 98th Annual Meeting (TRB)}, 2019.

\bibitem{Chao2020Survey}
Q.~Chao, H.~Bi, W.~Li, T.~Mao, Z.~Wang, M.~C. Lin, and Z.~Deng, ``A survey on visual traffic simulation: Models, evaluations, and applications in autonomous driving,'' \emph{Computer Graphics Forum}, vol.~39, no.~1, pp. 287--308, 2020.

\bibitem{Poudel2021Attack}
B.~Poudel and W.~Li, ``Black-box adversarial attacks on network-wide multi-step traffic state prediction models,'' in \emph{IEEE International Conference on Intelligent Transportation Systems (ITSC)}, 2021, pp. 3652--3658.

\bibitem{Lin2022GCGRNN}
L.~Lin, W.~Li, and L.~Zhu, ``Data-driven graph filter based graph convolutional neural network approach for network-level multi-step traffic prediction,'' \emph{Sustainability}, vol.~14, no.~24, p. 16701, 2022.

\bibitem{Guo2024Simulation}
K.~Guo, Z.~Miao, W.~Jing, W.~Liu, W.~Li, D.~Hao, and J.~Pan, ``Lasil: Learner-aware supervised imitation learning for long-term microscopic traffic simulation,'' in \emph{IEEE/CVF Conference on Computer Vision and Pattern Recognition (CVPR)}, 2024, pp. 15\,386--15\,395.

\bibitem{Poudel2022Micro}
B.~Poudel, T.~Watson, and W.~Li, ``Learning to control dc motor for micromobility in real time with reinforcement learning,'' in \emph{IEEE International Conference on Intelligent Transportation Systems (ITSC)}, 2022, pp. 1248--1254.

\bibitem{Wickman2022SparRL}
R.~Wickman, X.~Zhang, and W.~Li, ``A generic graph sparsification framework using deep reinforcement learning,'' in \emph{IEEE International Conference on Data Mining (ICDM)}, 2022, pp. 1221--1226.

\end{thebibliography}

\end{document}